\documentclass[letterpaper, 10 pt, conference]{ieeeconf}  

\IEEEoverridecommandlockouts                              

\usepackage{graphics} 
\usepackage{epsfig} 
\usepackage{mathptmx} 
\usepackage{times} 
\usepackage{amsmath} 
\usepackage{amssymb}  
\usepackage[caption=false,font=normalsize,labelfont=sf,textfont=sf]{subfig}
\usepackage{fixltx2e}
\usepackage{multirow}
\usepackage{dblfloatfix}
\DeclareMathOperator*{\argmax}{arg\,max}
\usepackage{algorithm}
\usepackage{algpseudocode}

\title{\LARGE \bf
Fragile object transportation by a multi-robot system in an unknown environment using a semi-decentralized control approach
}

\author{Dibyendu Roy$^{1}$, Sreejeet Maity$^{2}$, Madhubanti Maitra$^{3}$, and Samar Bhattacharya$^{3}$
\thanks{$^{1}$D. Roy is with TCS Research, India.
        {\tt\small roy.dibyendu@tcs.com}}%
\thanks{$^{2}$S. Maity is with Indian Institute of Science, Bangalore, India.
        }
\thanks{$^{3}$M. Maitra, and S. Bhattacharya are with Jadavpur University, Kolkata, India.
        }
}

\begin{document}

\maketitle
\thispagestyle{empty}
\pagestyle{empty}

\begin{abstract}

In this paper, we introduce a semi-decentralized control technique for a swarm of robots transporting a fragile object to a destination in an uncertain occluded environment. The proposed approach has been split into two parts. The initial part (Phase 1) includes a centralized control strategy for creating a specific formation among the agents so that the object to be transported, can be positioned properly on the top of the system. We present a novel triangle packing scheme fused with a circular region-based shape control method for creating a rigid configuration among the robots. 
In the later part (Phase 2), the swarm system is required to convey the object to the destination in a decentralized way employing the region based shape control approach. The simulation result as well as the comparison study demonstrates the effectiveness of our proposed scheme.

\end{abstract}

\section{INTRODUCTION}

A multi-robot system (MRS) is a group of robotic units that collaborate to offer affordable, fault-tolerant, and reliable solutions for a variety of automated applications \cite{chen2013dual}, \cite{gao2018review}, \cite{guo2019ultra}. 
One such application where an MRS could be employed to reduce human effort while providing flexibility, resilience, and a time-efficient solution is autonomous object transportation \cite{huzaefa2019centralized}, \cite{koung2021cooperative}, \cite{gombo2021communication}, \cite{tuci2018cooperative}.
Applying such a system to real-world tasks like waste retrieval and disposal, de-mining, collaborative structure construction, or object manipulation in an unknown environment where direct human intervention is impractical or impossible (such as in space or under the deep sea) could have socio-economic benefits.
Moreover, the MRS's decentralized feature enables physical separation and autonomous actions, leading to collective dexterity that a single robot, no matter how sophisticated or powerful, can't achieve.
These qualities are especially crucial in cooperative transport tasks \cite{tuci2018cooperative}, \cite{habibi2015distributed}, where the independent exertion of various pushing or pulling forces at different points of an object allows the group to produce precise maneuvers to avoid obstructions during transportation.

Most of the cooperative object transportation literature comprises multi-robot object carrying systems \cite{koung2021cooperative}-\cite{hawley2019control}, which solely rely on pushing or rolling mechanisms of a rigid geometrically shaped object. 
In \cite{ ardakani2017transporting}, a swarm of omnidirectional mobile robots employs frictional force to push and convey an elastic plate in an unknown environment. Similarly, \cite{hawley2019control} involves humanoid robots that introduce a central control scheme to exploit the dynamic stability of robots.  
A hierarchical quadratic programming-based optimization technique is presented in \cite{koung2021cooperative} for pushing and rolling a rigid object. This specific approach is a formation control problem. 
A communication-free decentralized solution has been proposed in \cite{ gombo2021communication} to transport flexible objects utilizing local force measurements. In this method, the swarm behavior is agent-dependent, obviating the need for a team structure.
The distributive motion planning scheme \cite{ habibi2015distributed} assists the swarm to compute the centroid of the object to be carried. As a consequence, the speed of the agents with respect to the object's centroid can be suitably modified for desired transit.

The aforementioned methods are categorized as formation control problems or pushing/rolling schemes of rigid objects. 
However, to carry a fragile object, the unit-loading scheme \cite{karabegovic2015application}, \cite{miller2017robots}, \cite{medina2020robotic} is preferable. It is a 
practical but less discussed problem, which needs to be resolved from an industrial standpoint. 
In this framework, the load is placed on top of the system so that it gets transported to the intended destination.
Reinforcement learning and sliding mode control based unit-loading schemes are presented in \cite{huzaefa2019centralized} and \cite{manko2018adaptive}
for cooperative rigid object transportation with a small number of robots. Since these techniques are not scalable in terms of the number of robots hence, heavy and fragile objects could not be carried by employing these strategies.

\begin{figure}[thpb]
\centering
\includegraphics[height=2in, width=\columnwidth]{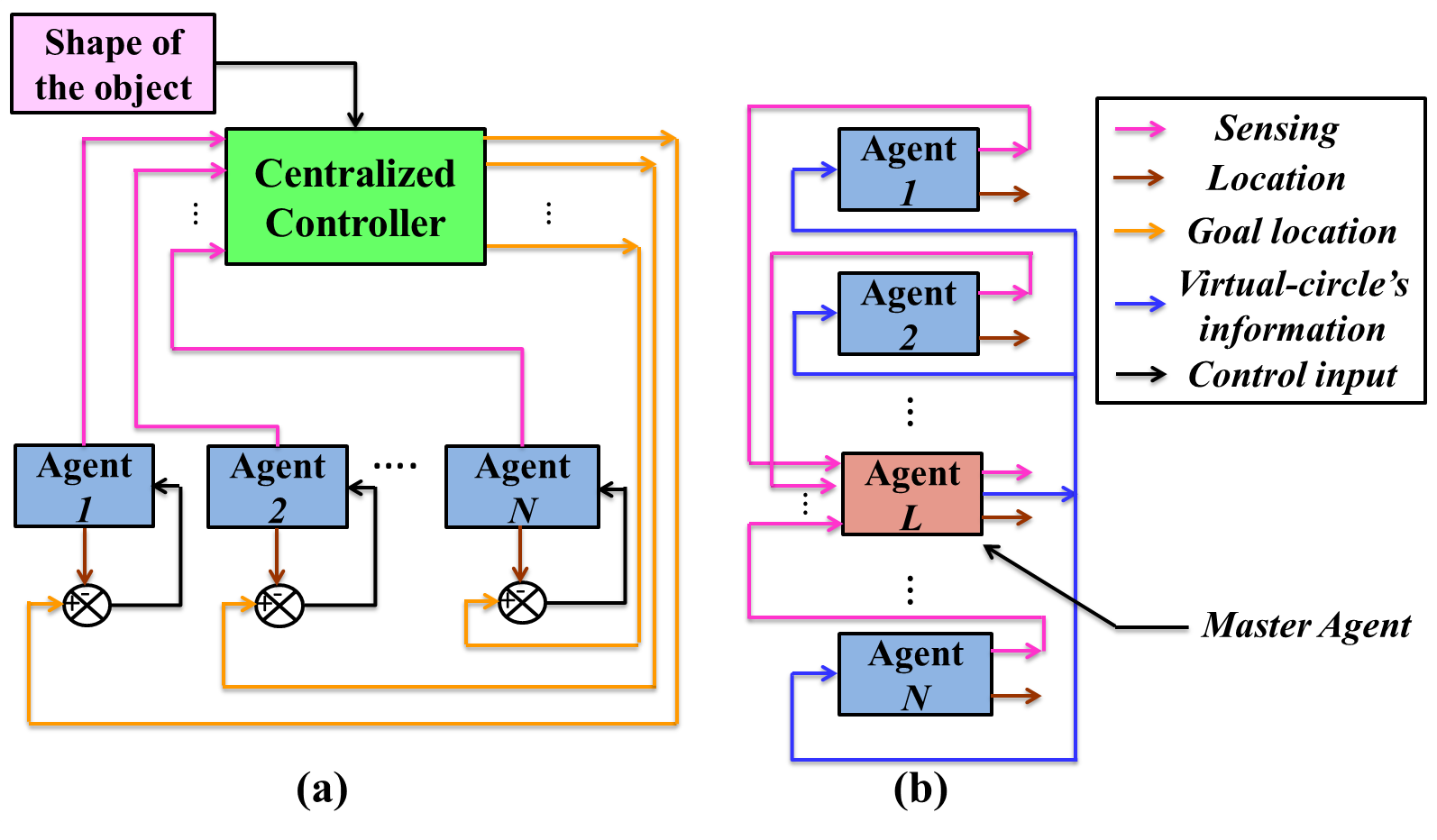}
\caption{Block diagram of the proposed control approach: (a) centralized control scheme in Phase-1, (b) decentralized control framework in Phase-2.}
\label{f1}
\end{figure}

To address the above issues, we propose a semi-decentralized control approach (the block diagram shown in Fig. \ref{f1}) for a swarm of robots to handle a unit-loading system. 
The strategy comprises two distinct phases. In Phase 1, a central controller is deployed, enabling arbitrarily placed agents to achieve a formation reliant on the shape of the object to be carried. Once the desired inter-agent formation is achieved, the object is loaded on top of the system.
After that, in Phase 2, the object is transported to a predefined destination in a decentralized fashion. 
In this work, we have only considered omnidirectional translation to minimize the severity of rotational impacts on fragile objects as the rotation might add unwanted crevasses on fine objects such as metal foils, papers, and brittle elements, thus, making the system well-suited for industrial applications.




The remainder of the paper is organized as follows. In Sec. II, the dynamics of the robotic system have been introduced. The centralized control system in Phase 1 has been discussed in Sec. III. Sec. IV illustrates the decentralized control architecture in Phase 2. The results and discussions are provided in Sec. V. Finally, Sec. VI concludes our work.

\section{DYNAMICS OF THE ROBOTIC SYSTEM}
Consider a group of $N$ robotic agents, operating in a 2-Dimensional (2D) space, that is required to carry an object towards a predefined target $T\in \Re^2$. The motion dynamics of the $i^{th}$ agent, positioned at $p^i\in\Re^2$, is defined as \cite{gazi2004class}
\begin{equation}
\dot{p}^i = -k_c(p^i-\Omega^i)+k_r\sum_{j=1,j\neq i}^N\left[\exp\left(-\frac{||p^{ij}||}{r_s^2}\right)(p^i-p^j) \right]
\label{e1}
\end{equation}
where, $\Omega^i\in\Re^2$ is an arbitrary point in the 2D space towards which the $i^{th}$ agent is required to approach, $k_c$ is the convergence gain, $||p^{ij}||=||p^i-p^j||$ is the Euclidean distance between the agents
$i$ and $j$, $r_s$ is the repulsion region, $k_r$ is the repulsion gain to avoid the inter-agent collision.
So, under the actuation of the above control law (Eq. \ref{e1}), the $i^{th}$ agent will approach towards $\Omega^i$ while avoiding inter-agent collision from its neighboring agents.

To sense the nearby obstacles, each agent is equipped with $m$ number of distance sensors \cite{roy2019virtual} with a spacing of $\frac{2\pi}{m}$ radian from each other. So, at the $t^{th}$ time, the sensed distance $(d^i(t))$ of obstacles around the $i^{th}$ agent becomes
\begin{equation}
d^i(t) = \begin{bmatrix}
d^i_1(t) & d^i_2(t) & ... & d^i_m(t) 
\end{bmatrix}^T
\label{e2}
\end{equation}
So, the predicted location coordinate of a sensed obstacle $\left(\hat{R}^i_j(t)=\begin{bmatrix}
\hat{R}^i_{j,x}(t) & \hat{R}^i_{j,y}(t)
\end{bmatrix}^T\right)$ from the $j^{th}$ sensor $(j\in m)$ with respect to the world coordinate frame of reference (WCF) would be
\begin{equation}
\hat{R}^i_j(t) = \begin{bmatrix}
\hat{R}^i_{j,x}(t) \\ \hat{R}^i_{j,y}(t)
\end{bmatrix} = \begin{bmatrix}
p^i_{x}(t) \\ p^i_{y}(t)
\end{bmatrix} + d^i_j \times
\begin{bmatrix}
\cos[(j-1)z+\beta^i(t)] \\ \sin[(j-1)z+\beta^i(t)]
\end{bmatrix}
\label{e3}
\end{equation}
where $p^i(t)=\begin{bmatrix}
p^i_{x}(t) & p^i_{y}(t)
\end{bmatrix}^T$ is the current position of the $i^{th}$ agent, $\beta^i(t)$ is the orientation of the agent with respect to WCF at the $t^{th}$ time, and $z = \frac{2\pi}{m}$.
Combining the predicted locations of all the sensors of $i^{th}$ agent, the consolidated locations of the nearby obstacles $(\hat{R}^i(t))$ would be
\begin{equation}
\hat{R}^i(t) = \begin{bmatrix}
\hat{R}^i_1(t) & \hat{R}^i_2(t) & ... & \hat{R}^i_m(t)
\end{bmatrix}\in\Re^{2\times m}
\label{e4}
\end{equation}
Now, combining the location coordinates of $N$ number of agents, the entire sensed zone of the swarm at the $t^{th}$ time will be computed as
\begin{equation}
\hat{R}(t) = \begin{bmatrix}
\hat{R}^1(t)^T & \hat{R}^2(t)^T & ... & \hat{R}^N(t)
\end{bmatrix}\in\Re^{m\times 2N}
\label{e4}
\end{equation}
Now, the question is how to compute the arbitrary points $\Omega=\{\Omega^1,\Omega^2,...,\Omega^N\}$ for $N$ number of agents (in each iteration) so that an object can be placed on the top of the system and carried afterward?
To solve this issue, in this paper, we divide the challenges into two phases; Phase 1 describes the precise estimation of $\Omega$ (under the assistance of a centralized controller) for loading the object on top of each agent. The second phase (Phase 2) evolves with each agent planning its path with respect to its neighboring agents to carry the object to the target. The step-by-step procedure is explained below.

\section{CONTROL SYSTEM IN PHASE 1}
In this phase, the agents must assemble into a precise formation in order to effectively load the object on top of each agent. A centralized controller has been employed to achieve this objective. 
The role of this controller is to reliably predict the set of arbitrary points $\Omega$, thus, acting as a supervisory controller. After getting this information, each agent will approach its target to create the formation.
The steps for finding $\Omega$ are outlined below.

\subsection{Fitting in virtual circular region}
The initial task of the central controller is to fit in a virtual circular region ($O_c(t)$) \cite{roy2018multi} having a radius of $r_c$ in the sensed space ahead. In this regard, we have taken two assumptions.

\textbf{Assumption 1 :} The diameter ($2r_{c}$) of the circular contour ($O_c(t)$) should be greater than or equal to the maximum dimension of the carrying object so that the object can be placed inside the virtual region.

\textbf{Assumption 2 :} There exist a path in the environment that allows the system to fit in circular regions of radius $r_c$.

Based on the predicted location coordinates of the nearby obstacles ($\hat{R}(t)$), the central controller will virtually fit in a circular region ($O_c(t)$) (having a radius of $r_c$) at the $t^{th}$ time. The detailed procedure of the circle fitting technique can be found in our previous paper \cite{roy2020geometric}.   

After evaluating the virtual circular region, the central controller will now compute the locations of the agents inside it for loading the object.
The best way to achieve this goal is to divide the circular region into small identical segments \cite{wang2009edge}. In this work, we employ the triangle packing concept \cite{wu2011mixed}, which produces a number of identical triangles to fill the virtual contour.
On the other hand, fitting identical triangles into a circle is an NP-hard problem \cite{chou2016np} that rarely yields optimal outcomes. Therefore, to solve this issue, initially, a square will be fitted inside the circle, followed by packing identical triangles inside the square as given below.

\subsection{Fitting square inside circle}
\begin{figure}[t]
\centering
\includegraphics[height=2.2in, width=3.5in]{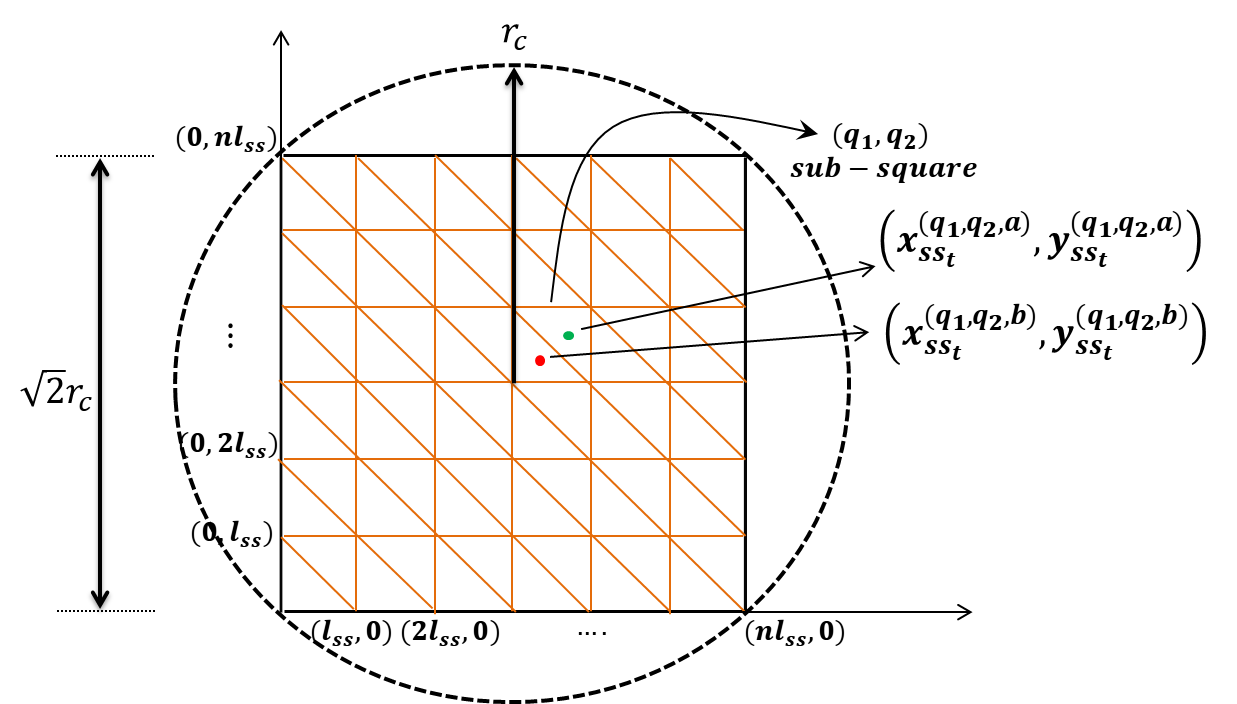}
\caption{The outcome of the triangle packing scheme within the circular region as defined in Phase-1.}
\label{f3}
\end{figure}

To fit in a square inside the circular region having a center at $O_c(t)=(x_c,y_c)$ and a radius of $r_c$, the length of each side of the square should be $\sqrt{2}r_c$. So, the coordinates of the four edge points of the square would be
\begin{equation}
\begin{bmatrix}
x_s \\ y_s
\end{bmatrix} = 
\begin{bmatrix}
x_c \\ y_c
\end{bmatrix} + r_c\times \sum_{k=1}^4 \begin{bmatrix}
\cos((k-1)\frac{\pi}{2}+\frac{\pi}{4}) \\ \sin((k-1)\frac{\pi}{2}+\frac{\pi}{4})
\end{bmatrix}
\label{e7}
\end{equation}
Once, the square is fitted inside the circle (as depicted in Fig. \ref{f3}), the next goal of the central controller is to pack the square with identical 
triangles.

\subsection{Packing triangles inside square}
To pack identical triangles into the fitted square (Fig. \ref{f3}), the simplest method is to divide the parent square into $n^2$ sub-squares. The sub-squares can easily be divided into equivalent triangles by connecting each sub-diagonal line segment. 
As a result, $2n^2$ number of right-angled isosceles triangles will be packed inside the parent square.

To divide the parent square having a side-length of $\sqrt{2}r_c$ into $n^2$ number of sub-squares, the length of each side of the sub-square would be $l_{ss}=\frac{\sqrt{2}r_c}{n}$ (Fig. \ref{f3}). 
So, the corner points of the $(q_1,q_2)^{th}$ sub-square $\left((q_1\times q_2) \in n^2\right)$ become
\begin{equation}
\begin{split}
\begin{bmatrix}
x^{(q_1,q_2)}_{ss} \\ y^{(q_1,q_2)}_{ss}
\end{bmatrix} &= 
\begin{bmatrix}
x_c \\ y_c
\end{bmatrix}+ \\ &\begin{bmatrix}
(q_2-1)l_{ss} & q_2l_{ss} & q_2l_{ss} & (q_2-1)l_{ss} \\ (q_1-1)l_{ss} & (q_1-1)l_{ss} & q_1l_{ss} & q_1l_{ss}
\end{bmatrix} 
\label{e8}
\end{split}
\end{equation}
After determining the $(q_1,q_2)^{th}$ sub-square, the next goal is to divide it into two identical triangles (presuming $(q_1,q_2,a)^{th}$, and $(q_1,q_2,b)^{th}$) by connecting the diagonal, as shown the Fig. \ref{f3}. So, the location coordinates of the centroids of those two right-angled isosceles triangles would be
\begin{equation}
\begin{bmatrix}
x^{(q_1,q_2,a)}_{ss_t} \\ y^{(q_1,q_2,a)}_{ss_t}
\end{bmatrix} = 
\begin{bmatrix}
x_c \\ y_c
\end{bmatrix}+ \\ \frac{1}{3}\times\begin{bmatrix}
(q_2-1)l_{ss}+q_2l_{ss}+(q_2-1)l_{ss} \\ (q_1-1)l_{ss}+(q_1-1)l_{ss}+q_1l_{ss}
\end{bmatrix} 
\label{e9.1}
\end{equation}
\begin{equation}
\begin{bmatrix}
x^{(q_1,q_2,b)}_{ss_t} \\ y^{(q_1,q_2,b)}_{ss_t}
\end{bmatrix} = 
\begin{bmatrix}
x_c \\ y_c
\end{bmatrix}+ \\ \frac{1}{3}\times\begin{bmatrix}
q_2l_{ss}+q_2l_{ss}+(q_2-1)l_{ss} \\ (q_1-1)l_{ss}+q_1l_{ss}+q_1l_{ss}
\end{bmatrix} 
\label{e9.2}
\end{equation}
Now, let us imagine that two agents are approaching toward the respective centroids of triangles $(q_1,q_2,a)$ and $(q_1,q_2,b)$. The Euclidean distance between the centroids should be greater than the repulsion-region $2r_s$ (Eq. \ref{e1}) for the agents to navigate steadily \cite{gazi2004class}. To accomplish this, the following theorem must be taken into account.

\textbf{Theorem 1:} For a stable navigation of the swarm, the number of sub-squares ($n$) along the horizontal or vertical axis of the parent square should be computed according to the following rule.
\[ n\leq \frac{r_c}{3\times r_s}\]

\textbf{Proof:} The proof is given in the supplementary document.

The next task of the central controller is to locate the $N$ number of triangular centroids (for $N$ number of agents) that should overlap with the shape of the object to be carried. This specific point is highlighted in the following section. 


\begin{figure}[thbp]
\centering
\includegraphics[height=2.5in, width=3.5in]{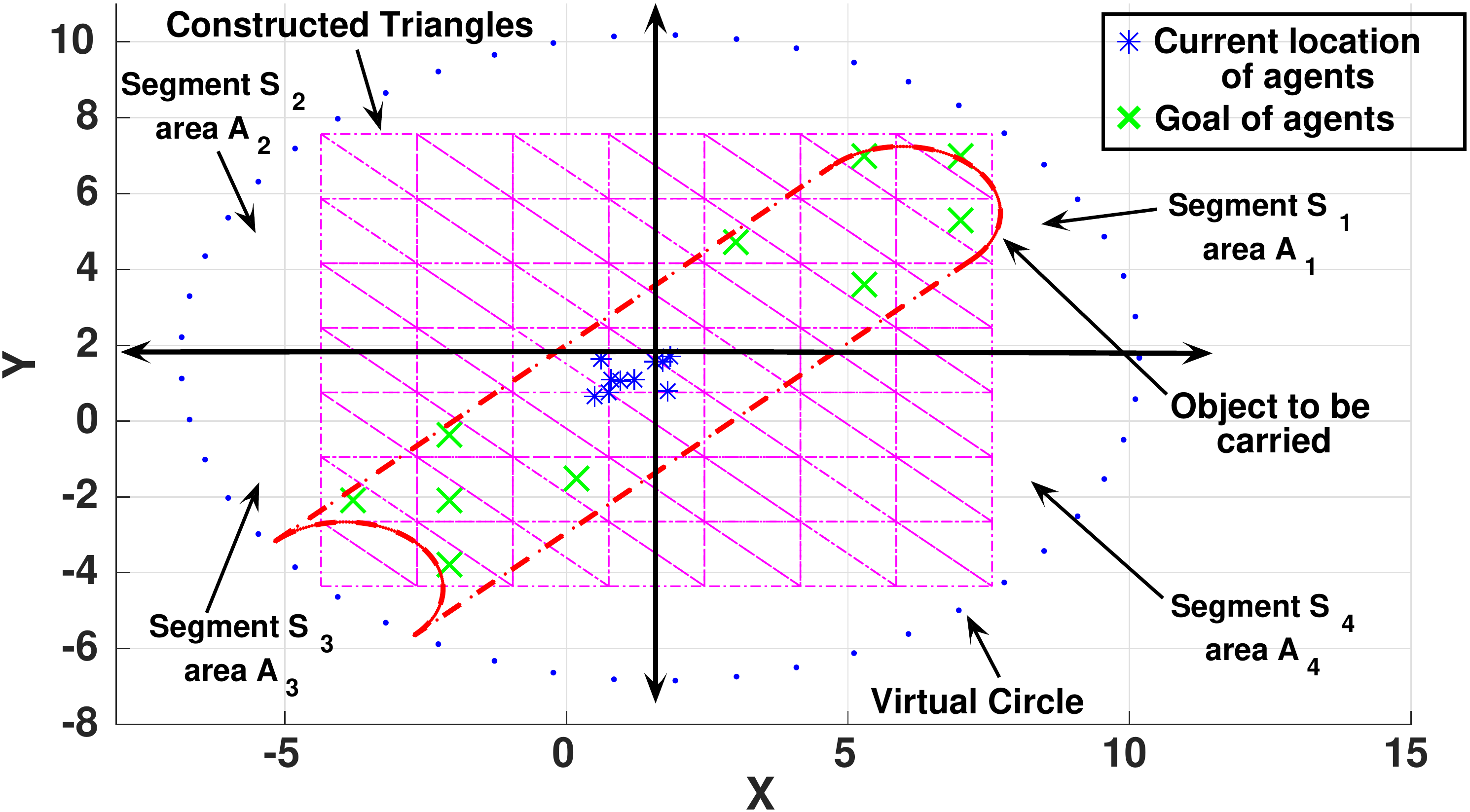}
\caption{Virtual segmentation of the circular region to determine the goal locations of the agents.}
\label{f4}
\end{figure}

\subsection{Finding goal-location of each agent}
In this step, in order to position the object on top of each agent, the central controller needs to precisely select $N$ points from the locations of the triangle centroids.
In this aspect, we assume that the controller is aware of the shape and size of the object. According to this consideration, the controller virtually splits the circular region into four segments (as illustrated in Fig. \ref{f4}), namely $S_1, S_2, S_3,$  and $S_4$ having areas of $A_1, A_2, A_3,$  and $A_4$ respectively such that $A_1=A_2=A_3=A_4=\frac{\pi r_c^2}{4}$.
Let us assume that the area of the object is $\Gamma$. The central controller uses the following equation to determine the number of triangles present in the $i^{th}$ segment $S_i;\forall i\in\{1,2,3,4\}$.
\begin{equation}
\begin{split}
\Delta_i=&\text{No. of Triangles within the area} (A_i\cap\Gamma) \\ &\forall i\in\{1,2,3,4\}
\end{split}
\label{e10}
\end{equation} 
After computing $\Delta_i$ of all the segments, the controller needs to select two segments which are having $\max\{\Delta_i\};\forall i=\{1,2,3,4\}$.

For the sake of simplicity, we assume that segments $S_1$, and $S_3$ are chosen as they have the most triangles (Fig. \ref{f4}). The locations of the centroids of those two segments would be in the following set:
\begin{equation}
Z^1_{ss_t}=\Big\{z_{ss_t}^{(1,1)}, z_{ss_t}^{(1,2)}, ..., z_{ss_t}^{(1,\Delta_1)}\Big\} 
\label{e11.1}
\end{equation}
\begin{equation}
Z^3_{ss_t}=\Big\{z_{ss_t}^{(3,1)}, z_{ss_t}^{(3,2)}, ..., z_{ss_t}^{(3,\Delta_3)}\Big\} 
\label{e11.2}
\end{equation}
where, $z_{ss_t}^{(1,w_1)}\in\Re^2;\forall w_1\in\{1,2,..,\Delta_1\}$, and $z_{ss_t}^{(3,w_2)}\in\Re^2;\forall w_2\in\{1,2,..,\Delta_3\}$ respectively.

Now, from the Eqs. \ref{e11.1}, and \ref{e11.2}, half of the goal coordinates (i.e., $N/2$) have been selected from the set $Z^1_{ss_t}$ and the remaining half will be chosen from set $Z^3_{ss_t}$, so that, the load of the object is evenly distributed between the two segments.
To select the goal coordinate of the $i^{th}$ agent from segment 1, the following goal-assignment strategy has been computed
\begin{equation}
\begin{split}
\Omega^i(t)=&\Big\{z_{ss_t}^{(1,w_1)} \mid \argmax_{z_{ss_t}^{(1,w_1)}\in Z_{ss_t}^1}||z^{(1,w_1)}_{ss_t}-O_c(t)||\Big\} \\ &\forall w_1\in\{1,2,...,\Delta_1\}; \quad \forall i\in\{1,2,...,N/2\}
\end{split}
\label{e11}
\end{equation}
From Eq. \ref{e11}, we can infer that agent 1 is assigned the centroid, which is farthest from the circular center, and the same procedure is followed for the other agents as well. 
Similarly, the goal-assignment strategy of the $j^{th}$ agent from segment 3 at the $t^{th}$ time could be defined as
\begin{equation}
\begin{split}
&\Omega^j(t)=\Big\{z_{ss_t}^{(3,w_2)} \mid \argmax_{z_{ss_t}^{(3,w_2)}\in Z_{ss_t}^3}||z^{(3,w_2)}_{ss_t}-O_c(t)||\Big\} \\ &\forall w_2\in\{1,2,...,\Delta_3\}; \quad \forall i\in\left\{\frac{N}{2}+1,...,N\right\}
\end{split}
\label{e11_1}
\end{equation}
As a result of Eqs. \ref{e11} and \ref{e11_1}, $N$ number of goal coordinates are evaluated for $N$ agents of the swarm. Fig. \ref{f4} displays one such outcome (marked by green crosses).
\begin{figure}[thbp]
\centering
\includegraphics[height=2.5in, width=3in]{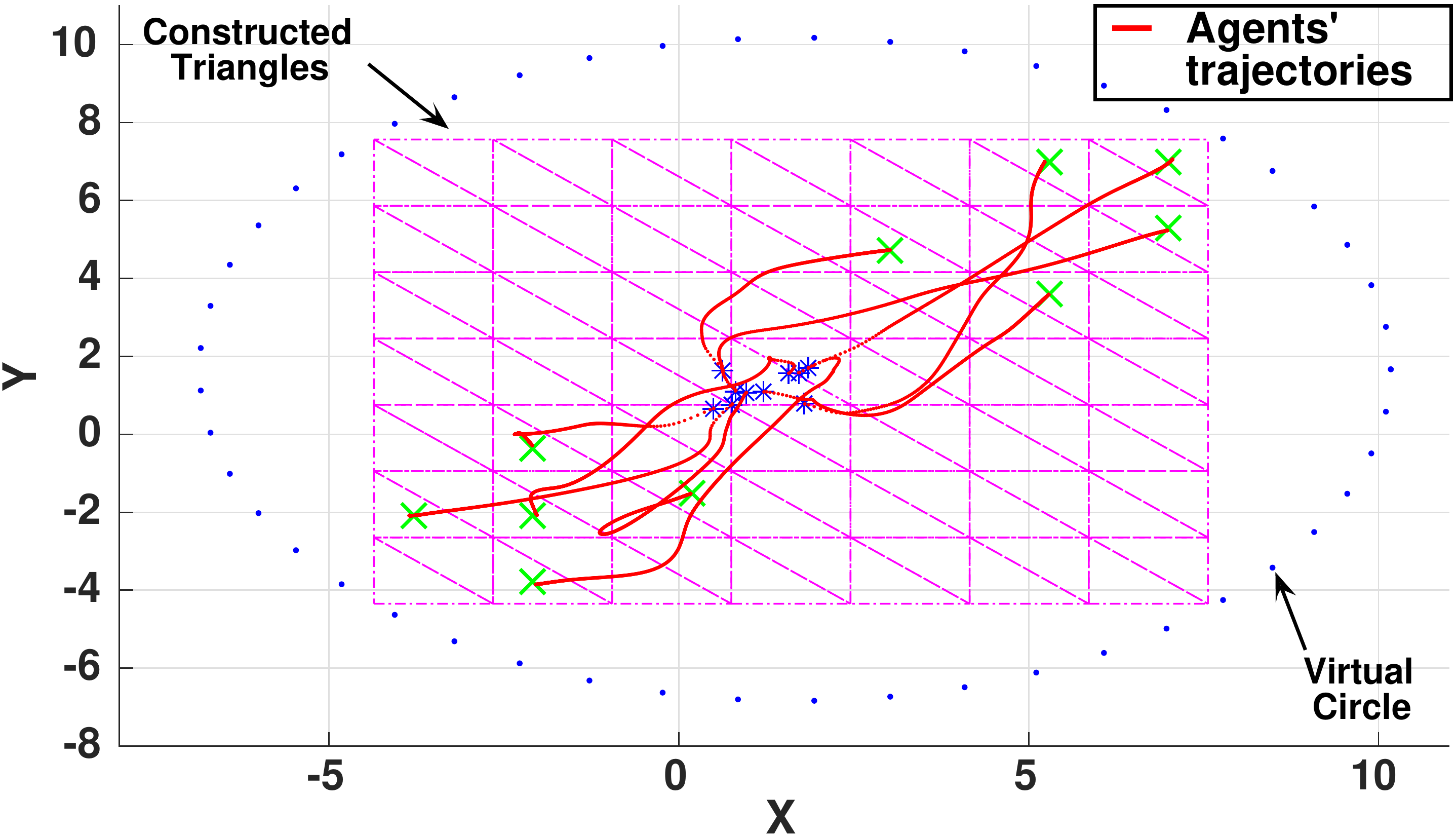}
\caption{Agents' trajectories after receiving the goal coordinate information from the central controller.}
\label{f5}
\end{figure}

After computing the goals of all the agents in the swarm, the central controller notifies every agent. Upon acquiring the goal location information, each agent will approach this point under the actuation of the controller as defined in Eq. \ref{e1}. The trajectories of all the agents are shown in Fig. \ref{f5}.


The duties of the central controller finish here. Before terminating the operation, the controller will pick one random agent from the swarm to serve as the \textit{master}-agent (Fig. \ref{f1}(b)). The task of this agent is to compute virtual circular regions in each time step of Phase 2 using the sensory data of all the agents (as discussed in Sec. III-A). Thus, the \textit{master}-agent will assist the swarm to navigate in Phase 2.

\begin{figure}[thbp]
\centering
\includegraphics[height=2.75in, width=3.3in]{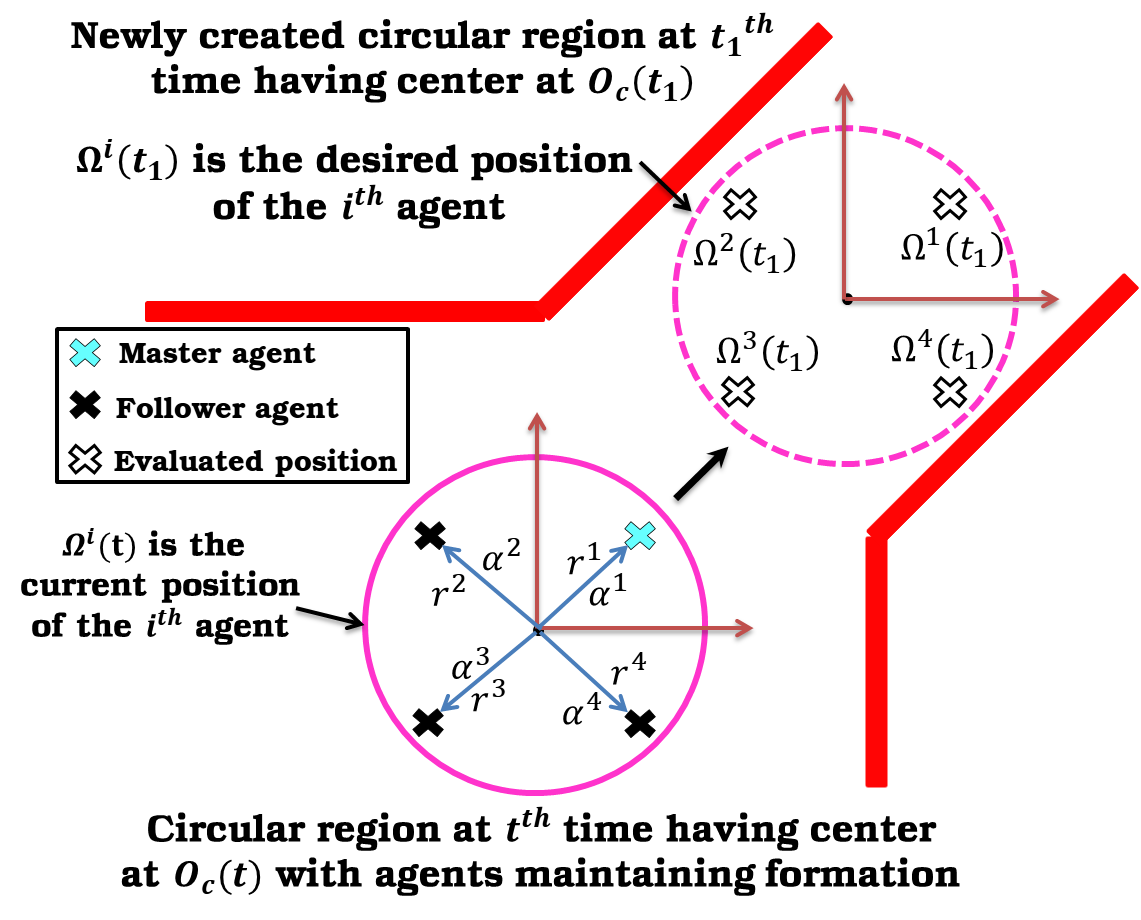}
\caption{Decentralized navigational strategy of Phase 2.}
\label{flc}
\end{figure}

\section{CONTROL SYSTEM IN PHASE 2}
In this phase of operation, each agent in the swarm will evaluate the location of its goal coordinate within the circular contour as specified by the \textit{master}-agent. Thus, the central controller is no longer required. 
As a consequence, the path planning of the swarm is assessed in a decentralized fashion in Phase 2.
In this work, we assume that the wheels of each agent are omnidirectional, hence, the swarm can only travel in one direction. Furthermore, a slight imbalance generated by the rotational movement of the system may break the object due to its fragility. 
Therefore, here, we only consider the fact that the orientation of the object is fixed during navigation to the destination.

In Phase 2, at the beginning of each iteration, all the agents in the swarm share their sensory information with the \textit{master}-agent (Fig. \ref{f1}(b)). After collecting all of the data, the \textit{master}-agent evaluates an analogous virtual circular region (having a radius of $r_c$) ahead.
Let us assume that the center coordinate of the newly created circular region is $O_c(t_1)=(x_{c_1},y_{c_1})$ (such that $t_1>t$) as presented in Fig. \ref{flc}. This information is then shared with all the agents. 
Now, the responsibility of each agent, including the \textit{master}-agent, is to compute the coordinate of its goal location within the newly formed circular contour $O_c(t_1)$ as explained in the upcoming section.

\begin{figure*}[]
\centering
\includegraphics[height=4in, width=6in]{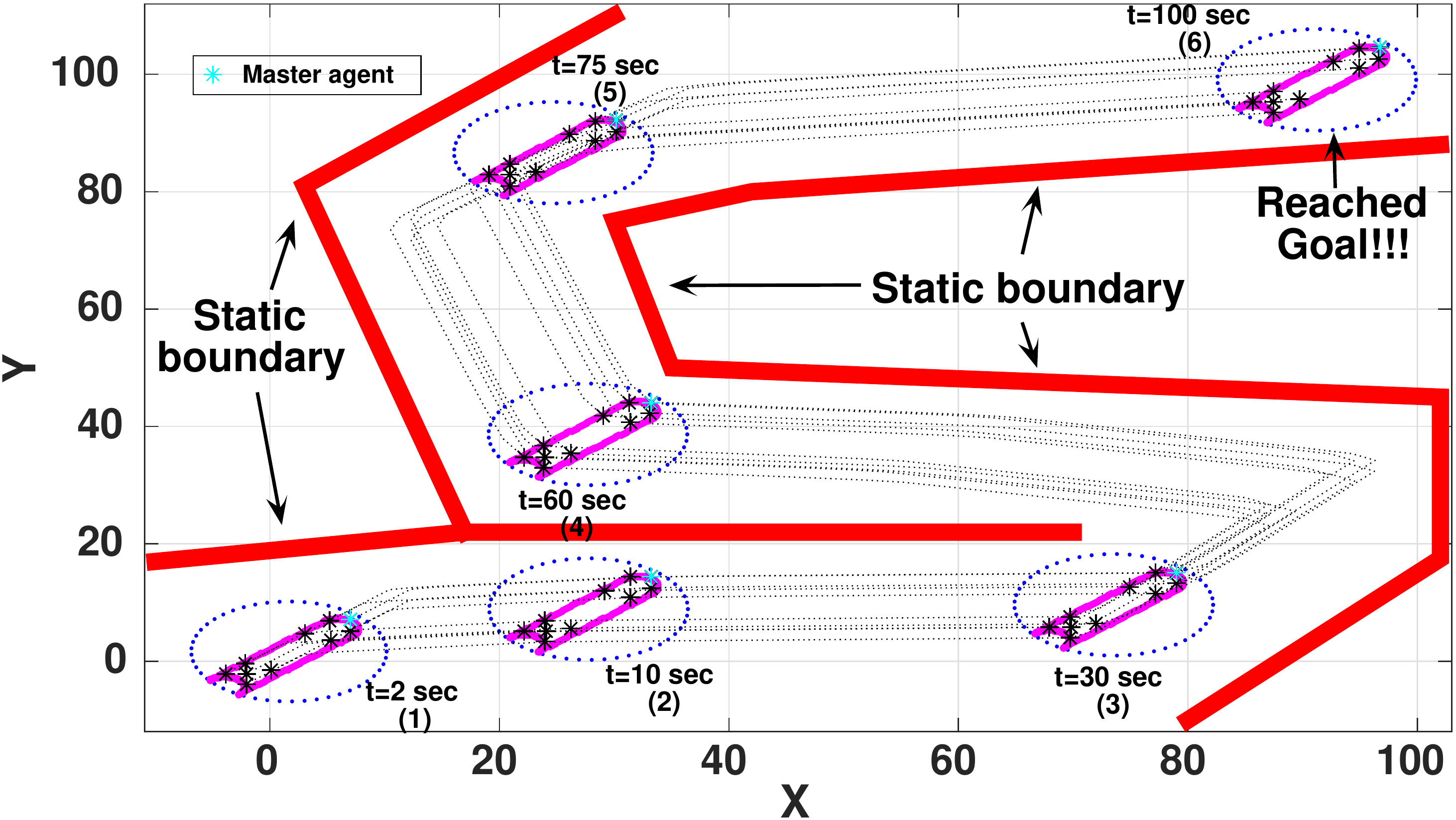}
\caption{Trajectories of all the agents (of Phase 2) carrying the object towards the destination using the proposed control technique.}
\label{f7}
\end{figure*}

\subsection{Evaluating goal coordinates within the circular region}
To determine the goal coordinate within the upcoming circular region, the $i^{th}$ agent first evaluates the distance and orientation of its current goal location ($\Omega^i(t)$) with respect to the present circular center $O_c(t)$ (Fig. \ref{flc}).

Let us assume that $r^i$ and $\alpha^i$ are the distance and orientation of the $i^{th}$ agent's current goal position ($\Omega^i(t)$) from the circular center $O_c(t)$ respectively (Fig. \ref{flc}). Hence, $r^i$ and $\alpha^i$ can be written as

\begin{equation}
r^i = ||\Omega^i(t)-O_c(t)||; \quad
\alpha^i = \tan^{-1}\left(\frac{\Omega^i(t)}{O_c(t)}\right)
\label{e15}
\end{equation}

Depending on the values of $r^i$ and $\alpha^i$, the upcoming goal location $\left(\Omega^i(t_1)=\begin{bmatrix}
\Omega^i_x(t_1) & \Omega^i_y(t_1)
\end{bmatrix}^T\right)$ of the $i^{th}$ agent with respect to the center coordinate ($O_c(t_1)$) of the newly created virtual region at the $t_1^{th}$ time could be computed as
\begin{equation}
\Omega^i(t_1) = \begin{bmatrix}
\Omega^i_x(t_1) \\ \Omega^i_y(t_1)
\end{bmatrix} = \begin{bmatrix}
x_{c_1} \\ y_{c_1}
\end{bmatrix} + r^i\times\begin{bmatrix}
\cos(\alpha^i) \\ \sin(\alpha^i)
\end{bmatrix}
\label{e16}
\end{equation}
where, $O_c(t_1)=(x_{c_1},y_{c_1})$ is the center-coordinate of the newly created circle. As the $i^{th}$ agent is an arbitrary agent, hence, the same holds true for all of the agents in the swarm. 
Moreover, considering the current ($\Omega^i(t)$) and future ($\Omega^i(t_1)$) goal locations of the $i^{th}$ agent, we can perceive that
\begin{equation}
||\Omega^i(t_1) - \Omega^i(t)|| = ||O_c(t_1) - O_c(t)|| \quad \forall i\in N
\label{lc}
\end{equation}

Hence, considering Eq. \ref{lc}, it is claimed that while moving from one circular region to another, each agent in the swarm will always travel an identical distance, which is solely dependent on the present and future center coordinates of the circular regions.
As a consequence, during navigation, all the agents would eventually approach their goals while maintaining uniform velocities. Thus, the proposed solution will reduce the risk of breaking the overhead object because of motion inconsistencies. 

%

The upcoming section presents the results and discussions of the proposed scheme.

\section{RESULTS AND DISCUSSIONS}
This section provides the simulation results to demonstrate the performance of the proposed control scheme in an occluded environment with narrow corners as shown in Fig. \ref{f7}. 
In Phase 1, a group of 10 robots initially placed in arbitrary scattered positions (indicated by blue stars of Fig. \ref{f5}) are required to form a desired formation under the influence of the centralized controller. Once the formation has been realized (Fig. \ref{f5}), the object is placed on top of the agents. Now in Phase 2, the agents are required to carry the object to the target location while circumventing barriers in a decentralized fashion. The simulation is performed on the MATLAB 2015a software platform with an Intel core-i7 processor having 16 GB of RAM. In this simulation study, the following parametric values have been chosen (Table I). 

\begin{table}[]
\centering
\caption{Values of the parameters}
\vspace{-0.3cm}
\begin{tabular}{|c|c|c|c|}
\hline
\textbf{Parameter} & \textbf{Value} & \textbf{Parameter} & \textbf{Value} \\ \hline
$k_c$              & 5              & $r_c$              & $8.5$          \\ \hline
$k_r$              & 2.5            & $T$                & {[}100, 100{]} \\ \hline
$r_s$              & 0.0575          & $n$                & 49             \\ \hline
$m$                & 8              & $N$                & 10             \\ \hline
\end{tabular}
\label{t1}
\end{table} 

In Phase 1, based on the agents' sensing input, the central controller creates a virtual circular region (blue dots in Fig. \ref{f5}). Then, several right-angled isosceles triangles (magenta lines of Fig. \ref{f5}) are packed into that area. After that, the central controller applies the goal-assignment strategy to estimate the goal locations of the 10 agents.
The object (shown in magenta contour of Fig. \ref{f7}) is loaded on top of the system ((1) of Fig. \ref{f7}) after all the agents have arrived their destinations (Fig. \ref{f5}). Phase 2 then begins, during which every agent in the swarm is in charge of path planning. In \ref{f7}, the \textit{master}-agent is highlighted in cyan-star. In this phase, the \textit{master}-agent evaluates the virtual circular regions (marked by blue-dot of Fig. \ref{f7}) in each iteration, which will assist the system to approach the target iteratively with the object on top while avoiding environmental impediments ((2), (3), (4), and (5) of Fig. \ref{f7}).
Finally, at $t=100$ sec, the system and the object have reached their destination ((6) of Fig. \ref{f7}).

\begin{figure}[]
\centering
\includegraphics[height=2.5in, width=3.2in]{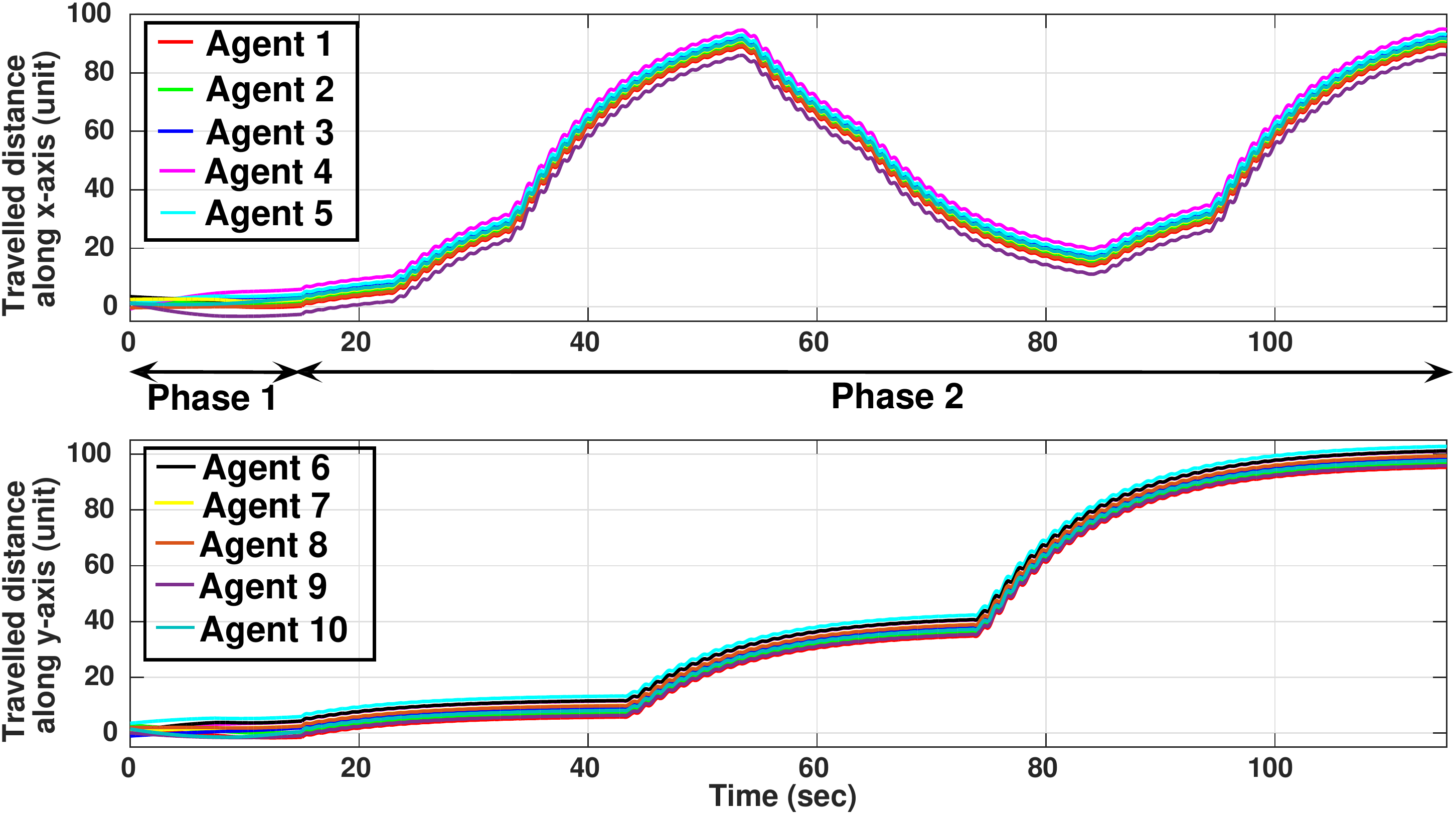}
\caption{Distance travel by all the agents during the navigation.}
\label{f8}
\end{figure}

Computing the total distance traveled by the agents (Fig. \ref{f8}), it is observed that Phase 1 consumes around 15 sec to execute. In Phase 2, all of the agents in the swarm maintain a uniform velocity since the distances between their present positions and the subsequent goal positions are fixed for all of them (Fig. \ref{flc}). The system needs 115 sec to reach at the target (Fig. \ref{f8}), so, Phase 2 takes 100 sec to finish the mission. 

\begin{figure}[]
\centering
\includegraphics[height=2in, width=3.2in]{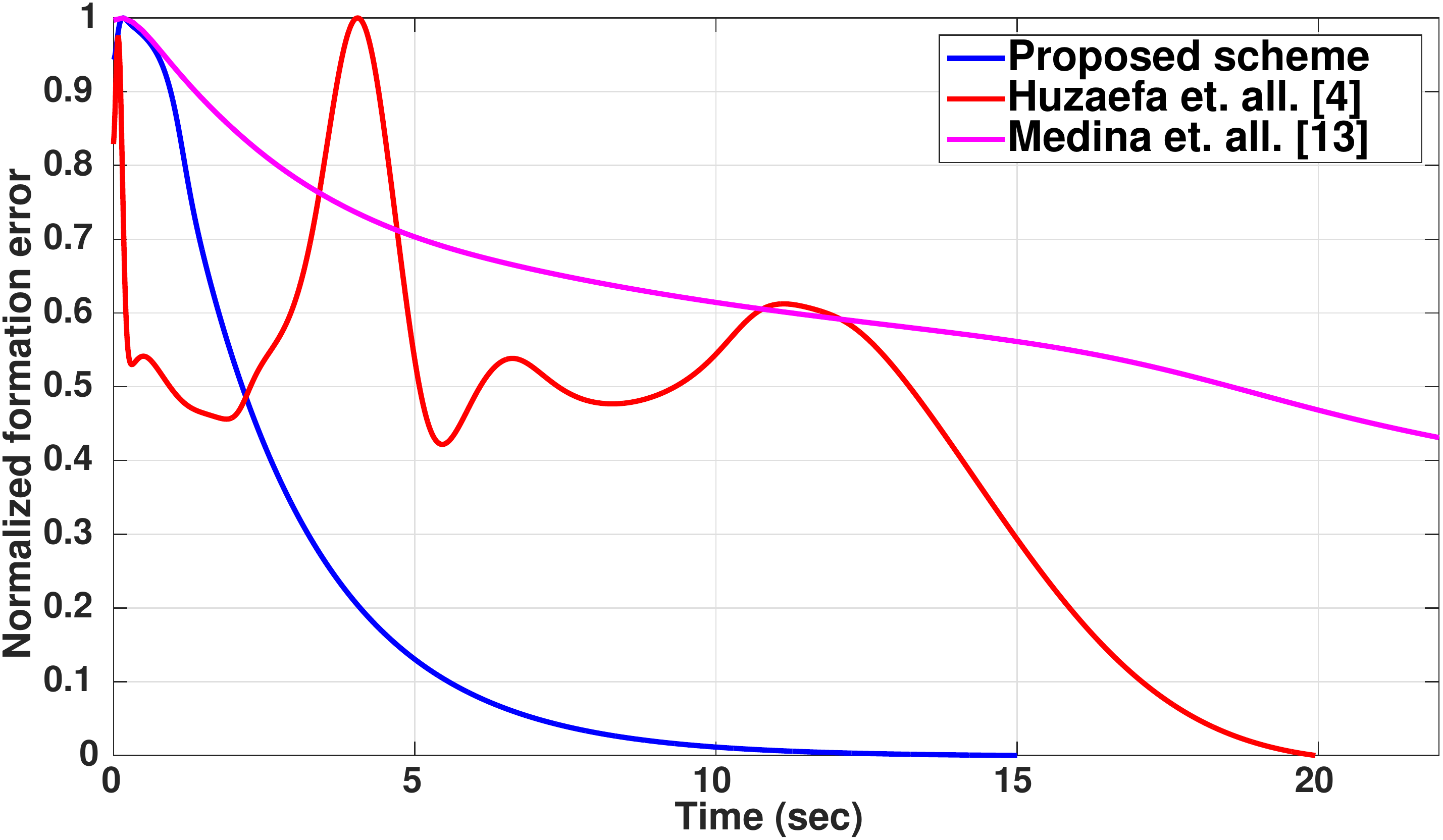}
\caption{Formation error comparison of the robotic swarm in Phase 1.}
\label{f9}
\end{figure}

We have executed another simulation study with the proposed technique for a non-convex environment with 10 robotic agents in the swarm. Here, the shape of the object is hexagonal.
The supplementary material contains the simulation result (Fig. SM.1).

\subsection{Comparison Analysis}
In this section, we would compare our proposed approach to some other well-cited literatures \cite{huzaefa2019centralized}, \cite{medina2020robotic} with respect to the formation error \cite{roy2021exploration} in Phase 1. The outcome is shown in Fig. \ref{f9}. The proposed semi-decentralized control approach consumes around 15 sec to achieve the desired configuration with 10 agents in the swarm. On the other hand, the centralized object transportation technique \cite{huzaefa2019centralized} takes around 20 sec to acquire the desired shape with 3 omnidirectional agents in the team. For communicating with the central controller during the entire navigational steps, this specific method consumes a significant amount of time. Finally, the decentralized object transportation scheme \cite{medina2020robotic} with 16 agents takes around 100 sec to attain the desired formation. 
In this case, all the agents communicate with their neighboring agents before executing any task, resulting in a significant delay. Thus, we can conclude from the comparative analysis that our proposed semi-decentralized control strategy outperforms the other unit-loading schemes.

\begin{table}[]
\center
\label{t2}
\caption{Algorithm convergence times comparison with scalable agents}
\begin{tabular}{|c|ccc|}
\hline
\multirow{2}{*}{}                 & \multicolumn{3}{c|}{\textbf{No. of agents}}                                      \\ \cline{2-4} 
                                  & \multicolumn{1}{c|}{\textbf{3}} & \multicolumn{1}{c|}{\textbf{7}}  & \textbf{10} \\ \hline
\textbf{Huzaefa et. all {[}4{]}}  & \multicolumn{1}{c|}{20-25 sec}  & \multicolumn{1}{c|}{78-92 sec}   & 124-145 sec \\ \hline
\textbf{Medina et. all. {[}13{]}} & \multicolumn{1}{c|}{49-62 sec}  & \multicolumn{1}{c|}{110-123 sec} & 152-180 sec \\ \hline
\textbf{Proposed}                 & \multicolumn{1}{c|}{26-35 sec}  & \multicolumn{1}{c|}{55-62 sec}   & 91-100 sec  \\ \hline
\end{tabular}
\end{table}

In Table II, we compare the algorithm convergence times of our proposed approach with the centralized \cite{huzaefa2019centralized} and decentralized \cite{medina2020robotic} control schemes with the scalability of agents in a similar environment as presented in Fig. SM.1 of the supplementary material. We observe that the centralized approach \cite{huzaefa2019centralized} performs well when there are few agents in the swarm. However, with scalable agents, both centralized and decentralized strategies perform unsatisfactorily than the offered approach.

Hence, from the above results and comparisons, we can conclude that the semi-decentralized control approach might be an alternative for transporting a fragile object from the source to the destination by a swarm of robots.

\section{CONCLUSIONS}
In this paper, we have proposed a semi-decentralized control technique for a swarm of robots to convey an object to a destination in an unknown environment. 
In the initial part (Phase 1), a centralized controller has been employed to build a strict inter-agent formation to place the object on top of the swarm system. A triangle packing scheme integrated with the region-based shaped control approach has been introduced to form the firm inter-agent formation amongst the agents in the swarm. 
In the later part (Phase 2), the swarm carries the object to the desired target in a decentralized fashion by preserving a similar inter-agent pattern. To illustrate the performance of the proposed controller, simulation results have been presented. In the future, we would like to use this technique in a real environment with physical robots.




%


\bibliographystyle{IEEEtran}
\bibliography{IEEEabrv,dib}

\end{document}